\documentclass[10pt,twocolumn,letterpaper]{article}

\usepackage{cvpr}              

\usepackage{graphicx}
\usepackage{amsmath}
\usepackage{amssymb}
\usepackage{booktabs}
\usepackage{times}
\usepackage{epsfig}
\usepackage{multirow}
\usepackage{mathtools}
\usepackage{xcolor}
\usepackage{flexisym}
\usepackage{adjustbox}

\def\etal{\emph{et al.}}
\def\eg{\emph{e.g.}}

\def\wrt{\emph{w.r.t.}}

\definecolor{cfgreen}{rgb}{0.0, 0.42, 0.24}

\DeclarePairedDelimiter{\norm}{\lVert}{\rVert}

\usepackage[pagebackref,breaklinks,colorlinks]{hyperref}

\usepackage[capitalize]{cleveref}
\crefname{section}{Sec.}{Secs.}
\Crefname{section}{Section}{Sections}
\Crefname{table}{Table}{Tables}
\crefname{table}{Tab.}{Tabs.}

\def\cvprPaperID{7738} 
\def\confName{CVPR}
\def\confYear{2022}

\begin{document}

\title{ Sparse to Dense Dynamic 3D Facial Expression Generation}



\author{Naima Otberdout\\
Univ. Lille, CNRS, Inria, Centrale Lille, UMR 9189 CRIStAL, F-59000 Lille, France\\
{\tt\small naima.otberdout@univ-lille.fr}
\and
Claudio Ferrari\\
Deparment of Architecture and Engineering \\
University of Parma, Italy\\
{\tt\small claudio.ferrari2@unipr.it}
\and
Mohamed Daoudi\\
IMT Nord Europe, Institut Mines-Télécom, Univ. Lille, Centre for Digital Systems, F-59000 Lille, France\\
Univ. Lille, CNRS, Centrale Lille, Institut Mines-Télécom, UMR 9189 CRIStAL, F-59000 Lille, France\\
{\tt\small mohamed.daoudi@imt-nord-europe.fr}
\and
Stefano Berretti\\
Media Integration ad Communication Center \\
University of Florence, Italy\\
{\tt\small stefano.berretti@unifi.it}
\and
Alberto Del Bimbo\\
Media Integration ad Communication Center \\
University of Florence, Italy\\
{\tt\small alberto.delbimbo@unifi.it}
}
\maketitle

\begin{abstract}
In this paper, we propose a solution to the task of generating dynamic 3D facial expressions from a neutral 3D face and an expression label. This involves solving two sub-problems: \textit{(i)} modeling the temporal dynamics of expressions, and \textit{(ii)} deforming the neutral mesh to obtain the expressive counterpart. We represent the temporal evolution of expressions using the motion of a sparse set of 3D landmarks that we learn to generate by training a manifold-valued GAN (Motion3DGAN). To better encode the expression-induced deformation and disentangle it from the identity information, the generated motion is represented as per-frame displacement from a neutral configuration. To generate the expressive meshes, we train a Sparse2Dense mesh Decoder (S2D-Dec) that maps the landmark displacements to a dense, per-vertex displacement. This allows us to learn how the motion of a sparse set of landmarks influences the deformation of the overall face surface, independently from the identity. Experimental results on the CoMA and D3DFACS datasets show that our solution brings significant improvements with respect to previous solutions in terms of both dynamic expression generation and mesh reconstruction, while retaining good generalization to unseen data. The code and the pretrained model will be made publicly available.
\end{abstract}


\section{Introduction}\label{sec:intro}
Synthesizing dynamic 3D (4D) facial expressions aims at generating realistic face instances with varying expressions or speech-related movements that dynamically evolve across time, starting from a face in neutral expression. It finds application in a wide range of graphics applications spanning from 3D face modeling, to augmented and virtual reality for animated films and computer games. While recent advances in generative neural networks have made possible the development of effective solutions that operate on 2D images~\cite{Fan-AAAI:2019, OtberdoutPAMI2020}, the literature on the problem of generating facial animation in 3D is still quite limited. 

\begin{figure}[!t]
\centering 
\includegraphics[width=\linewidth]{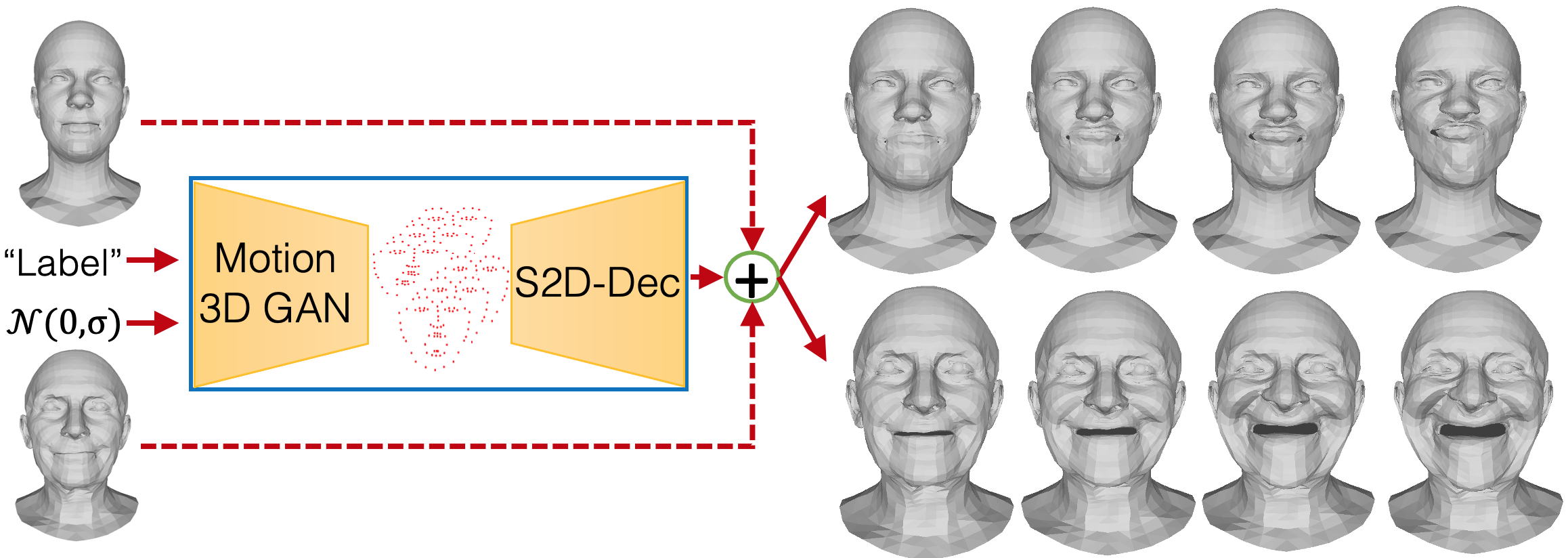}
\caption{\textbf{3D dynamic facial expression generation}: A GAN generates the motion of 3D landmarks from an expression label and noise; A decoder expands the animation from the landmarks to a dense mesh, while keeping the identity of a neutral 3D face,}
\label{Fig:overview}
\end{figure}

To perform a faithful and accurate 3D facial animation, three main challenges arise. First, the identity of the subject whose neutral face is used as starting point for the sequence should be maintained across time. Second, the applied deformation should correspond to the specified expression/motion that is provided as input, and should be applicable to any neutral 3D face. Incidentally, these are major challenges in 3D face modeling, which require disentangling structural face elements related to the identity, \eg, nose or jaw shape, from deformations related to the movable face parts, \eg, mouth opening/closing. 
Finally, it is required to model the temporal dynamics of the specified expression so to obtain realistic animations. 

Some previous works tackled the problem by capturing the facial expression of a subject frame-by-frame and transferring it to a target model~\cite{cao-tog:2014}. However, in this case the temporal evolution is not explicitly modeled, so the problem reduces to transferring a tracked expression to a neutral 3D face. Some other works animated a 3D face mesh given an arbitrary speech signal and a static 3D face mesh as input~\cite{Cudeiro-cvpr:2019, Karras-tog:2017}. Also in this case, the temporal evolution is guided by an external input, similar to a tracked expression. Instead, here we are interested in animating a face just starting from a neutral face and an expression label. 

In our solution, which is illustrated in Figure~\ref{Fig:overview}, the temporal evolution and the mesh deformation are decoupled and modeled separately in two network architectures. A manifold-valued GAN (\textit{Motion3DGAN}) accounts for the expression dynamics by generating a temporally consistent motion of 3D landmarks corresponding to the input label from noise. The landmarks motion is encoded using the \textit{Square Root Velocity Function} (SRVF) and compactly represented as a point on a hypersphere. Then, a Sparse2Dense mesh Decoder (\textit{S2D-Dec}) generates a dense 3D face guided by the landmarks motion for each frame of the sequence. To effectively disentangle identity and expression components, the landmarks motion is represented as a per-frame displacement from a neutral configuration. Instead of directly generating a mesh, the S2D-Dec expands the landmarks displacement to a dense, per-vertex displacement, which is finally used to deform the neutral mesh. The intuition that led to this architecture is the following: the movement induced on the face surface by the underlying facial muscles is consistent across subjects. In addition, it causes the vertex motion to be locally correlated as muscles are smooth surfaces. We thus train the decoder to learn how the displacement of a sparse set of points influences the displacement of the whole face surface. This has the advantage that structural face parts, \eg, nose or forehead, which are not influenced by facial expressions are not deformed, helping in maintaining the identity traits stable. Furthermore, the network can focus on learning expressions at a fine-grained level of detail and generalize to unseen identities.


In summary, the main contributions of our work are: 
\emph{(i)} we propose an original method to generate dynamic sequences of 3D expressive scans given a neutral 3D mesh and an expression label. Our approach has the capability of generating strong and diverse expression sequences, with high generalization ability to unseen identities and expressions; 
\emph{(ii)} we adapt a specific GAN architecture~\cite{OtberdoutPAMI2020} for dynamic 3D landmarks generation, and design a decoder for expressive mesh reconstruction from a neutral mesh and landmarks. Differently from common auto-encoders, the proposed S2D-Dec learns to generate a per-vertex displacement map from a few control points, allowing accurate mesh deformations where the structural face parts remain stable;
\emph{(iii)} we designed a novel reconstruction loss that weighs the contribution of each vertex based on its distance from the landmarks. This proved to augment the decoder capability of generating accurate expressions. 


\section{Related Work}\label{sect:related-work}
Our work is related to methods for 3D face modeling, facial expression generation guided by landmarks, and dynamic generation of 3D faces, \ie, 4D face generation. 

\textbf{3D face modeling}. 
The 3D Morphable face Model (3DMM) as originally proposed in~\cite{blanz1999morphable} is the most popular solution for modeling 3D faces. 
The original model and its variants~\cite{Booth-cvpr:2016, Paysan-avss:2009, brunton2014multilinear, neumann2013sparse, luthi2017gaussian, ferrari2021sparse, ferrari2015dictionary} capture face shape variations both for identity and expression based on linear formulations, thus incurring in limited modeling capabilities. For this reason, non-linear encoder-decoder architectures are attracting more and more attention. This comes at the cost of reformulating convolution and pooling/unpooling like operations on the non-regular mesh support~\cite{Bronstein17, Litany-cvpr:2018, Verma-cvpr:2018}. Ranjan~\etal~\cite{COMA:ECCV18} proposed an auto-encoder architecture that builds upon newly defined spectral convolution operators, and pooling operations to down-/up-sample the mesh. Bouritsas~\etal~\cite{Bouritsas:ICCV2019} improved upon the above by proposing a novel graph convolutional operator enforcing consistent local orderings on the vertices of the graph through the \textit{spiral operator}~\cite{lim2018_correspondence_learning}. Despite their impressive modeling precision, a recent work~\cite{ferrari2021sparse} showed that they heavily suffer from poor generalization to unseen identities. This limits their practical use in tasks such as face fitting or expression transfer. We finally mention that other approaches exist to learn generative 3D face models, such as~\cite{Abrevaya-wacv:2018, Moschoglou-ijcv:2020}. However, instead of dealing with meshes they use alternative representations for 3D data, such as depth images or UV-maps.

To overcome the above limitation, we go beyond self-reconstruction and propose a mesh decoder that, differently from previous models, learns expression-specific mesh deformations from a sparse set of landmark displacements.

\textbf{Facial expression generation guided by landmarks}.
Recent advances in neural networks made facial landmark detection reliable and accurate both in 2D~\cite{Chen-NEURIPS:2019, Dong-tpami:2020, Wan-TNNLS:2021} and 3D~\cite{gilani2017deep, zhu2017face}. Landmarks and their motion are a viable way to account for facial deformations as they reduce the complexity of the visual data, and have been commonly used in several 3D face related tasks, \eg, reconstruction~\cite{ferrari2015dictionary, FLAME:SiggraphAsia2017} or reenactment~\cite{ferrari2018rendering, garrido2014automatic}. Despite some effort was put in developing landmark-free solutions for 3D face modeling~\cite{chang2018expnet, chang2017faceposenet, gecer2019ganfit}, some recent works investigated their use to model the dynamics of expressions. Wang~\etal~\cite{WangCVPR2018} proposed a framework that decouples facial expression dynamics, encoded into landmarks, and face appearance using a conditional recurrent network. Otberdout~\etal~\cite{OtberdoutPAMI2020} proposed an approach for generating videos of the six basic expressions given a neutral face image. The geometry is captured by modeling the motion of landmarks with a GAN that learns the distribution of expression dynamics. 

These methods demonstrated the potential of using landmarks to model the dynamics of expressions and generate 2D videos. In our work, we instead tackle the problem of modeling the dynamics in 3D, exploring the use of the motion of 3D landmarks to both model the temporal evolution of expressions and animate a 3D face.

\textbf{4D face generation}. 
While many researchers tackled the problem of 3D mesh deformation, the task of 3D facial motion synthesis is yet more challenging. A few studies addressed this issue by exploiting audio features~\cite{Zeng-ACMMM:2020, Karras-tog:2017}, speech signal~\cite{Cudeiro-cvpr:2019} or tracked facial expressions~\cite{cao-tog:2014} to generate facial motions. However, none of these explicitly model the temporal dynamics and resort to external information. 

To the best of our knowledge, the work in~\cite{PotamiasECCV2020} is the only approach that specifically addressed the problem of dynamic 3D expression generation. In that framework, the motion dynamics is modeled with a temporal encoder based on an LSTM, which produces a per-frame latent code starting from a per-frame expression label. The codes are then fed to a mesh decoder that, similarly to our approach, generates a per-vertex displacement that is summed to a neutral 3D face to obtain the expressive meshes. Despite the promising results reported in~\cite{PotamiasECCV2020}, we identified some limitations.  First, the LSTM is deterministic and for a given label the exact same displacements are generated. Our solution instead achieves diversity in the output sequences by generating from noise. Moreover, in~\cite{PotamiasECCV2020} the mesh decoder generates the displacements from the latent codes, making it dependent from the temporal encoder. In our solution, the motion dynamics and mesh displacement generation are decoupled, using landmarks to link the two modules. The S2D-Dec is thus independent from Motion3DGAN, and can be used to generate static meshes as well given a arbitrary set of 3D landmarks as input. This permits us to use the decoder for other tasks such as expression/speech transfer. Finally, as pointed out in~\cite{PotamiasECCV2020}, the model cannot perform extreme variations well. Using landmarks allowed us to define a novel reconstruction loss that weighs the error of each vertex with respect to its distance from the landmarks, encouraging accurate modeling of the movable parts. Thanks to this, we are capable of accurately reproducing from slight to strong expressions, and generalize to unseen motions.

\section{Proposed Method}
\begin{figure*}[!ht]
\centering 
\includegraphics[width=0.9\linewidth]{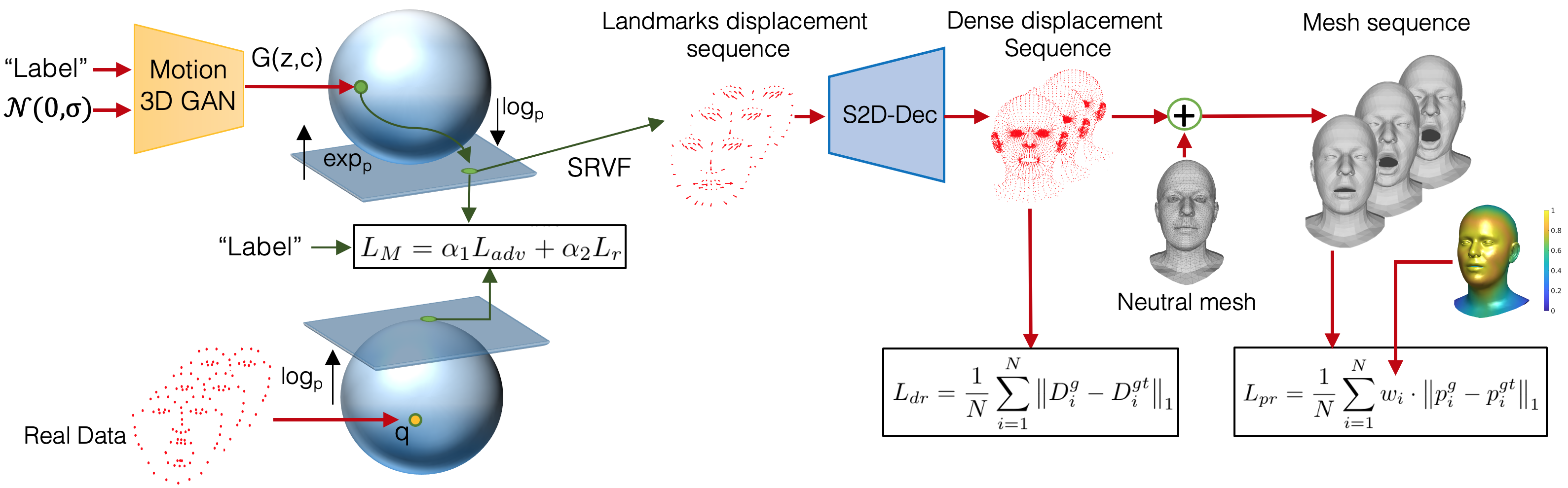}
\caption{\textbf{Overview of our framework.} Motion3DGAN generates the motion $q(t)$ of 3D landmarks corresponding to an expression label from a noise vector $z$. The module is trained guided by a reconstruction loss $L_r$ and adversarial loss $L_{adv}$. The motion $q(t)$ is converted to a sequence of landmark displacements $d_i$, which are fed to S2D-Dec. From each $d_i$, the decoder generates a dense displacement $D_i^g$. A neutral mesh is then summed to the dense displacements to generate the expressive meshes $\mathbf{S}^g$. S2D-Dec is trained under the guidance of a displacement loss $L_{dr}$ and our proposed weighted reconstruction loss $L_{pr}$.}
\label{fig:detailed-approach}
\end{figure*}

Our approach consists of two specialized networks as summarized in Figure~\ref{fig:detailed-approach}. 
Motion3DGAN accounts for the temporal dynamics and generates the motion of a sparse set of 3D landmarks from noise, provided an expression label, \eg, happy, angry. The motion is represented as per frame landmark displacements with respect to a neutral configuration. These displacements are fed to a decoder network, S2D-Dec, that constructs the dense point-cloud displacements from the sparse displacements given by the landmarks. These dense displacements are then added to a neutral 3D face to generate a sequence of expressive 3D faces corresponding to the initial expression label. 
In the following, we separately describe the two networks. 

\subsection{Generating Sparse Dynamic 3D Expressions}
Facial landmarks were shown to well encode the temporal evolution of facial expressions~\cite{kacem2017novel, OtberdoutPAMI2020}.  
Motivated by this fact, we generate the facial expression dynamics based on the motion of 3D facial landmarks. Given a set of $k$ 3D landmarks, $Z(t) = {(x_i(t),y_i(t),z_i(t))}_{i=1}^{k}$, with $Z(0)$ being the neutral configuration, their motion can be seen as a trajectory in $\mathbb{R}^{k \times 3}$ and can be formulated as a parameterized curve in $\mathbb{R}^{k \times 3}$ space. 
Let $\alpha: I = [0,1] \rightarrow \mathbb{R}^{ k \times 3}$ represent the parameterized curve, where each $\alpha(t) \in \mathbb{R}^{k \times 3}$. For the purpose of modeling and studying our curves, we adopt the Square-Root Velocity Function (SRVF) proposed in~\cite{SrivastavaKJJ11}. The SRVF $q(t): I \rightarrow \mathbb{R}^{k \times 3}$ is defined by:
%
\begin{equation}
\label{eq:curve_2_q}
q(t) = \frac{\dot{\alpha}(t)}{\sqrt{ \|\dot{\alpha}(t)\|_2}},
\end{equation}

\noindent
with the convention that $q(t) = 0$ if $\dot{\alpha}(t)=0$. 
This function proved effective for tasks such as human action recognition~\cite{devanne20143} or 3D face recognition~\cite{drira20133d}. Similar to this work, Otberdout~\etal~\cite{OtberdoutPAMI2020} proposed to use the SRVF representation to model the temporal evolution of 2D facial landmarks, which makes it possible to learn the distribution of these points and generate new 2D facial expression motions. 
In this paper, we extend this idea to 3D by proposing the Motion3DGAN model to generate the motion of 3D facial landmarks represented using the SRVF encoding in~\eqref{eq:curve_2_q}. 

Following~\cite{OtberdoutPAMI2020}, we remove the scale variability of the resulting motions by scaling the $\mathbb{L}^2$-norm of these functions to $1$. As a result, we transform the motion of 3D facial landmarks to points on a hypersphere  $\mathcal{C}=\{q:[0,1] \rightarrow \mathbb{R}^{k \times 3}, \|q\|=1\}$. 
The resulting representations are manifold-valued data that cannot be handled with traditional generative models. 

To learn the distribution of the SRVF representations, we propose Motion3DGAN as an extension of MotionGAN~\cite{OtberdoutPAMI2020}, a conditional version of the Wasserstein GAN for manifold-valued data~\cite{HuangAAAI19}. It maps a random vector $z$ to a point on the hypersphere $\mathcal{C}$ conditioned on an input class label. Motion3DGAN is composed of two networks trained adversarially: a generator $G$ that learns the distribution of the 3D landmark motions, and a discriminator $D$ that distinguishes between real and generated 3D landmark motions. Motion3DGAN is trained by a weighted sum of an adversarial loss $L_{adv}$ and a reconstruction loss $L_{r}$ such that $L_{M}= \alpha_1 L_{adv} + \alpha_2 L_{r}$. The former is given by:
\begin{equation}
\begin{array}{rl}
L_{adv}= & {\mathbb{E}_{q \sim \mathbb{P}_{q}}\left[D\left(\log _{p}(q),c\right)\right]} 
\\ 
{ } & {-\mathbb{E}_{z \sim \mathbb{P}_{z}}\left[D\left(\log _{p}\left(\exp _{p}(G(z, c))\right)\right)\right]}
\\ 
{ } & {+\lambda E_{\hat{q} \sim \mathbb{P}_{\hat{q}}}\left[\left(\left\|\nabla_{\hat{q}} D(\hat{q})\right\|_{2}-1\right)^{2}\right]}.
\end{array}
\label{Eq:LossWassersteinGAN}
\end{equation}

\noindent
In~\eqref{Eq:LossWassersteinGAN}, $q \sim \mathbb{P}_{q}$ is an SRVF sample from the training set, $c$ is the expression label (\eg, mouth open, eyebrow) that we encode as a one-hot vector and concatenate to a random noise $z \sim \mathbb{P}_{z}$. The last term of the adversarial loss represents the gradient penalty of the Wasserstein GAN~\cite{gulrajani2017improved}. 
Specifically, $\hat{q} \sim \mathbb{P}_{\hat{q}}$ is a random point sampled uniformly along straight lines between pairs of points sampled from $\mathbb{P}_{q}$ and the generated distribution $\mathbb{P}_{g}$:
\begin{equation}
\hat{q} = (1 - \tau) \log_p(q) + \tau\log_p(\exp_p(G(z, c))), 
\end{equation}

\noindent
where $0 \leq \tau \leq 1$, and $\nabla_{\hat{q}} D(\hat{q})$ is the gradient \wrt~$\hat{q}$. 
The functions $\log_p(.)$ and $\exp_p(.)$ are the logarithm and the exponential maps, respectively, defined in a particular point $p$ of the hypersphere. They map the SRVF data forth and back to a tangent space of $\mathcal{C}$ (details in the supplementary material).
%
%
Finally, the reconstruction loss is defined as:
\begin{equation}
\label{eq:T_reconstruction_loss}
L_{r} = \|{ \log_{p}(\exp_p({G(z,c)})) - \log_p(q)} \|_1,
\end{equation}

\noindent
where $\|.\|_1$, represents the $L_1$-norm, and $q$ is the ground truth SRVF corresponding to the condition $c$. The generator and discriminator architectures are similar to~\cite{OtberdoutPAMI2020}.

The SRVF representation is reversible, which makes it possible to recover the curve $\alpha(t)$ from a new generated SRVF $q(t)$ by,
\begin{equation}
\label{eq:curve}
\alpha(t) = \int_0^t \norm{q(s)}q(s)ds + \alpha(0).
\end{equation}

\noindent 
Where $\alpha(0)$ represents the initial landmarks configuration $Z(0)$. Accordingly, using this equation, we can apply the generated motion to \textit{any} landmark configuration, making it robust to identity changes.

\subsection{From Sparse to Dense 3D Facial Expressions}
Our final goal is to animate the neutral mesh $\mathbf{S}^{n}$ to obtain a novel 3D face $\mathbf{S}^{g}$ reproducing some expression, yet maintaining the identity structure of $\mathbf{S}^{n}$. 
Given this, we point at generating the displacements of the mesh vertices from the sparse displacements of the landmarks to animate $\mathbf{S}^{n}$. In the following, we assume all the meshes have a fixed topology, and are in full point-to-point correspondence. 

Let $\mathcal{L} = \left\{\left(\mathbf{S}^{n}_1, \mathbf{S}^{gt}_1, Z^n_1, Z^{gt}_1\right), \ldots,\left(\mathbf{S}^{n}_{m}, \mathbf{S}^{gt}_m, Z^{gt}_m, Z^n_m\right)\right\}$ be the training set, where $\mathbf{S}^{n}_i = (p^n_1, \ldots, p^n_N) \in \mathbb{R}^{N \times 3}$ is a neutral 3D face, $\mathbf{S}^{gt}_i = (p^{gt}_1, \ldots, p^{gt}_N) \in \mathbb{R}^{N \times 3}$ is a 3D expressive face, $Z^n_i \in \mathbb{R}^{k \times 3}$ and $Z^{gt}_i \in \mathbb{R}^{k \times 3}$ are the 3D landmarks corresponding to $\mathbf{S}^{n}_i$ and $\mathbf{S}^{gt}_i$, respectively. 
We transform this set to a training set of sparse and dense displacements, $\mathcal{L}\textprime = \left\{\left(D_1, d_1\right), \ldots, \left(D_{m}, d_m\right)\right\}$ such that, $D_i = \mathbf{S}^{gt}_i - \mathbf{S}^{n}_i$ and $d_i=Z^{gt}_i - Z^{n}_i$. Our goal here is to find a mapping $h: \mathbb{R}^{k \times 3} \rightarrow \mathbb{R}^{N \times 3}$ such that $\mathbf{D}_i \approx h\left(d_i\right)$.
We designed the function $h$ as a decoder network (S2D-Dec), where the mapping is between a sparse displacement of a set of landmarks and the dense displacement of the entire mesh points. Finally, in order to obtain the expressive mesh, the dense displacement map is summed to a 3D face in neutral expression, \ie, $\mathbf{S}^{e}_i = \mathbf{S}^{n}_i + \mathbf{D}_i$. 
The S2D-Dec network is based on the spiral operator proposed in~\cite{Bouritsas:ICCV2019}. Our architecture includes five spiral convolution layers, each one followed by an up-sampling layer. More details on the architecture can be found in the supplementary material.

In order to train this network, we propose to use two different losses, one acting directly on the displacements and the other controlling the generated mesh. The reconstruction loss of the dense displacements is given by,
\begin{equation}
\label{eq:displacement_loss}
L_{dr} = \frac{1}{N}\sum_{i=1}^{N}  \left \| D^{g}_i - D^{gt}_i \right \|_1,
\end{equation}

\noindent
where $D^{g}$ and $D^{gt}$ are the generated and the ground truth dense displacements, respectively. To further improve the reconstruction accuracy, we add a loss that minimizes the error between $\mathbf{S}^{g}$ and the ground truth expressive mesh $\mathbf{S}^{gt}$. We observed that vertices close to the landmarks are subject to stronger deformations. Other regions like the forehead, instead, are relatively stable. To give more importance to those regions, we defined a weighted version of the $L1$ loss:
\begin{equation}
\label{eq:L1_weighted}
L_{pr}= \frac{1}{N}\sum_{i=1}^{N} w_i \cdot \left \| p^{g}_i - p^{gt}_i \right \|_1 .
\end{equation}

\noindent
We define the weights as the inverse of the Euclidean distance of each vertex $p_i$ in the mesh from its closest landmark $Z_j$, \ie $w_i = \frac{1}{\min d(p_i, Z_j)}, \; \forall j$. This provides a coarse indication of how much each $p_i$ contributes to the expression generation. Since the mesh topology is fixed, we can pre-compute the weights $w_i$ and re-use them for each sample. Weights are then re-scaled so that they lie in $[0, 1]$. Vertices corresponding to the landmarks, \ie, $p_i = Z_j$ for some $j$, are hence assigned the maximum weight. 
We will show this strategy provides a significant improvement with respect to the standard $L1$ loss. The total loss used to train the S2D-Dec is given by $L_{S2D}=\beta_1. L_{dr} + \beta_2. L_{pr}$.

\section{Experiments}
We validated the proposed method in a broad set of experiments on two publicly available benchmark datasets.

\noindent
\textbf{CoMA dataset}~\cite{COMA:ECCV18}: It is a common benchmark employed in other studies~\cite{Bouritsas:ICCV2019, COMA:ECCV18}. It consists of $12$ subjects, each one performing $12$ extreme and asymmetric expressions. Each expression comes as a sequence of meshes $\mathbf{S} \in \mathbb{R}^{N \times 3}$ ($140$ meshes on average), with $N = 5,023$ vertices. \\
\textbf{D3DFACS dataset}~\cite{cosker2011facs}: We used the registered version of this dataset~\cite{li2017learning}, which has the same topology of CoMA. It contains $10$ subjects, each one performing a different number of facial expressions. In contrast to CoMA, this dataset is labeled with the activated action units of the performed facial expression. It is worthy to note that the expressions of D3DFACS are highly different from those in CoMA.

\subsection{Training Details}
In order to keep Motion3DGAN and S2D-Dec  decoupled, they are trained separately. We used CoMA to train Motion3DGAN, since this dataset is labeled with facial expression classes.  We manually divided each sequence into sub-sequences of length $30$ starting from the neutral to the apex frame. The first sub-sequence for each subject and expression is used as test set. We use all the others for training as we generate from a random noise at test time.\footnote{For reproducibility, the list of the sub-sequences used to train Motion3DGAN will be publicly released.}. Then, we encoded the motion of $k=68$ landmarks from the sub-sequences in the SRVF representation, and used them to train Motion3DGAN. The landmarks were first centered and normalized to unit norm. Each of the $12$ expression labels was encoded as a one-hot vector, concatenated with a random noise vector of size $128$.  


To comprehensively evaluate the capability of S2D-Dec of generalizing to either unseen identities or expressions, we performed subject-independent and expression-independent cross-validation experiments. For the subject-independent experiment, we used a $4$-fold cross-validation protocol for CoMA, training on $9$ and testing on $3$ identities in each fold. On D3DFACS, we used the last $7$ identities for training and the remaining $3$ as test set. Concerning the expression-independent splitting, we used a $4$-fold  cross-validation protocol for CoMA, training on $9$ and testing on $3$ expressions in each fold. For D3DFACS, given the different number of expression per subject, the first $11$ expressions were used for testing and trained on the rest.

We trained both Motion3DGAN and S2D-Dec using the Adam optimizer, with learning rate of $0.0001$ and $0.001$ and mini-batches of size $128$ and $16$, respectively. Motion3DGAN was trained for $8000$ epochs, while $300$ epochs were adopted for S2D-Dec. The hyper-parameters of the Motion3DGAN and S2D-Dec losses were set empirically to $\alpha_1 = 1$, $\alpha_2 = 10$, $\beta_1= 1$ and $\beta_2=0.1$. We chose the mean SRVF of the CoMA data as a reference point $p$, where we defined the tangent space of $\mathcal{C}$. 


\subsection{3D Expression Generation}
For evaluation, we set up a baseline by first comparing against standard 3DMM-based fitting methods. Similar to previous works~\cite{ferrari2017dictionary, FLAME:SiggraphAsia2017}, we fit $\mathbf{S}^n$ to the set of target landmarks $Z^{e}$ using the 3DMM components. Since the deformation is guided by the landmarks, we first need to select a corresponding set from $\mathbf{S}^n$ to be matched with $Z^{e}$. Given the fixed topology of the 3D faces, we can retrieve the landmark coordinates by indexing into the mesh, \ie, $Z^n = \mathbf{S}^n(\mathbf{I}_z)$, where $\mathbf{I}_z \in \mathbb{N}^{n}$ are the indices of the vertices that correspond to the landmarks. We then find the optimal deformation coefficients that minimize the Euclidean error between the target landmarks $Z^{e}$ and the neutral ones $Z^{n}$, and use the coefficients to deform $\mathbf{S}^n$. 
In the literature, 
several 3DMM variants have been proposed. We experimented the standard PCA-based 3DMM and the DL-3DMM in~\cite{ferrari2017dictionary}. We chose this latter variant as it is conceptually similar to our proposal, being constructed by learning a dictionary of deformation displacements. For fair comparison, we built the two 3DMMs using a number of deformation components comparable to the size of the S2D-Dec input, \ie, $68 \times 3 = 204$. For PCA, we used either $38$ components (retaining the $99\%$ of the variance) and $220$, while for DL-3DMM we used $220$ dictionary atoms.

With the goal of comparing against other deep models, we also considered the Neural3DMM~\cite{Bouritsas:ICCV2019}. It is a mesh auto-encoder tailored for learning a non-linear latent space of face variations and reconstructing the input 3D faces. In order to compare it with our model, we modified the architecture and trained the model to generate an expressive mesh $\mathbf{S}^g$ given its neutral counterpart as input. To do so, we concatenated the landmarks displacement (of size $204$) to the latent vector (of size $16$) and trained the network towards minimizing the same $L_{pr}$ loss used in our model. 
All the compared methods were trained on the same data. 
Finally, we also identified the FLAME model~\cite{FLAME:SiggraphAsia2017}. Unfortunately, the training code is not available, and using the model pre-trained on external data would not be a fair comparison. 

The mean per-vertex Euclidean error between the generated meshes and their ground truth is used as standard performance measure, as in the majority of works~\cite{Bouritsas:ICCV2019, COMA:ECCV18, ferrari2021sparse, PotamiasECCV2020}. Note that we exclude the Motion3DGAN model here as we do not have the corresponding ground-truth for the generated landmarks (they are generated from noise). Instead, we make use of the ground truth motion of the landmarks.

\subsubsection{Comparison with Other Approaches}
\begin{figure}[!t]
  \centering
    \includegraphics[width=\linewidth]{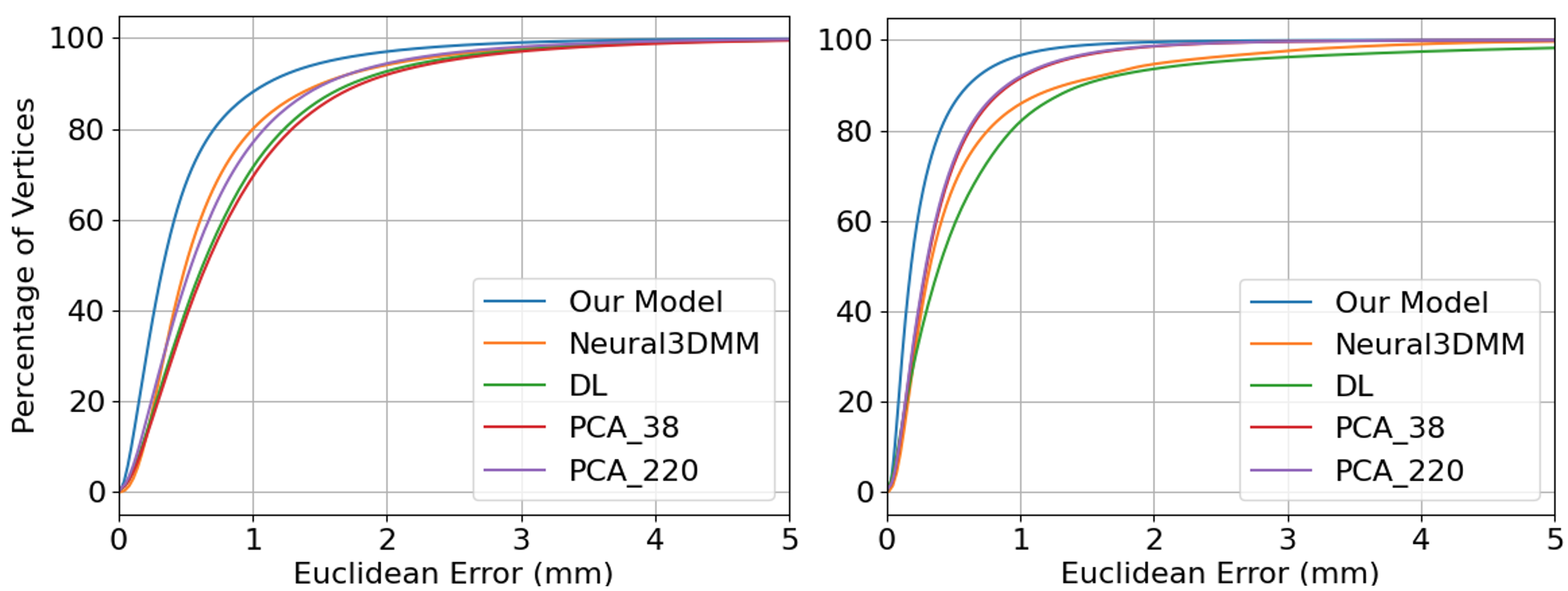}
   \caption{Cumulative per-vertex Euclidean error between PCA-based 3DMM models, DL-3DMM, Neural3DMM, and our proposed model, using expression-independent cross-validation on the CoMA (left) and D3DFACS (right) datasets.}
   \label{fig:cum-err-exprind}
\end{figure}

Table~\ref{tab:comparison-ExprSplit} shows a clear superiority of S2D-Dec over state-of-the-art methods for both the protocols and datasets, proving its ability to generate accurate expressive meshes close to the ground truth in both the case of unseen identities or expressions. In Figure~\ref{fig:cum-err-exprind}, the cumulative per-vertex error distribution on the expression-independent splitting further highlights the precision of our approach, which can reconstruct 90\%-98\% of the vertices with an error lower than $1mm$. While other fitting-based methods retain satisfactory precision in both the protocols, we note that the performance of Neural3DMM~\cite{Bouritsas:ICCV2019} significantly drop when unseen identities are considered. This outcome is consistent to that reported in~\cite{ferrari2021sparse}, in which the low generalization ability of these models is highlighted. We also note that results for the identity-independent protocol were never reported in the original papers~\cite{COMA:ECCV18, Bouritsas:ICCV2019}. Overall, our solution embraces the advantages of both approaches, being as general as fitting solutions yet more accurate.

\begin{table}[!t]
\centering
\small
\begin{adjustbox}{width=\columnwidth,center}
\begin{tabular}{@{}l@{}cc@{}|cc@{}}
\toprule
& \multicolumn{2}{c}{Expression Split} & \multicolumn{2}{c}{Identity Split}\\
\midrule
Method & CoMA& D3DFACS & CoMA & D3DFACS\\
\midrule
PCA-220 & $0.76 \pm 0.73$ & $0.42 \pm 0.44$ & $0.80 \pm 0.73$ & $0.56 \pm 0.56$\\
PCA-38 & $0.90 \pm 0.84$ & $0.44\pm0.45 $ & $0.93 \pm 0.82$ & $0.58 \pm 0.56$\\
DL3DMM~\cite{ferrari2017dictionary} & $0,86 \pm 0,80$  & $0.73\pm 1.15$ & $0.89 \pm 0.79$ & $1.15 \pm 1.50$\\
Neural~\cite{Bouritsas:ICCV2019} & $0.75 \pm 0.85$ & $0.59 \pm 0.86$ & $3.74 \pm 2.34$ & $2.09\pm 1.37$\\
\midrule
{\bf Ours} & $\mathbf{0.52 \pm 0.59}$ & $\mathbf{0.28\pm 0.31}$ & $\mathbf{0.55 \pm 0.62}$ & $\mathbf{0.27 \pm 0.30}$\\
\bottomrule
\end{tabular}
\end{adjustbox}
\caption{Reconstruction error (mm) on expression-independent (left) and identity-independent (right) splits: comparison with PCA-$k$ 3DMM ($k$ components), DL-3DMM (220 dictionary atoms), and Neural3DMM.}
\label{tab:comparison-ExprSplit}
\end{table}

\begin{figure}[!t]
\centering
\includegraphics[width=0.8\linewidth]{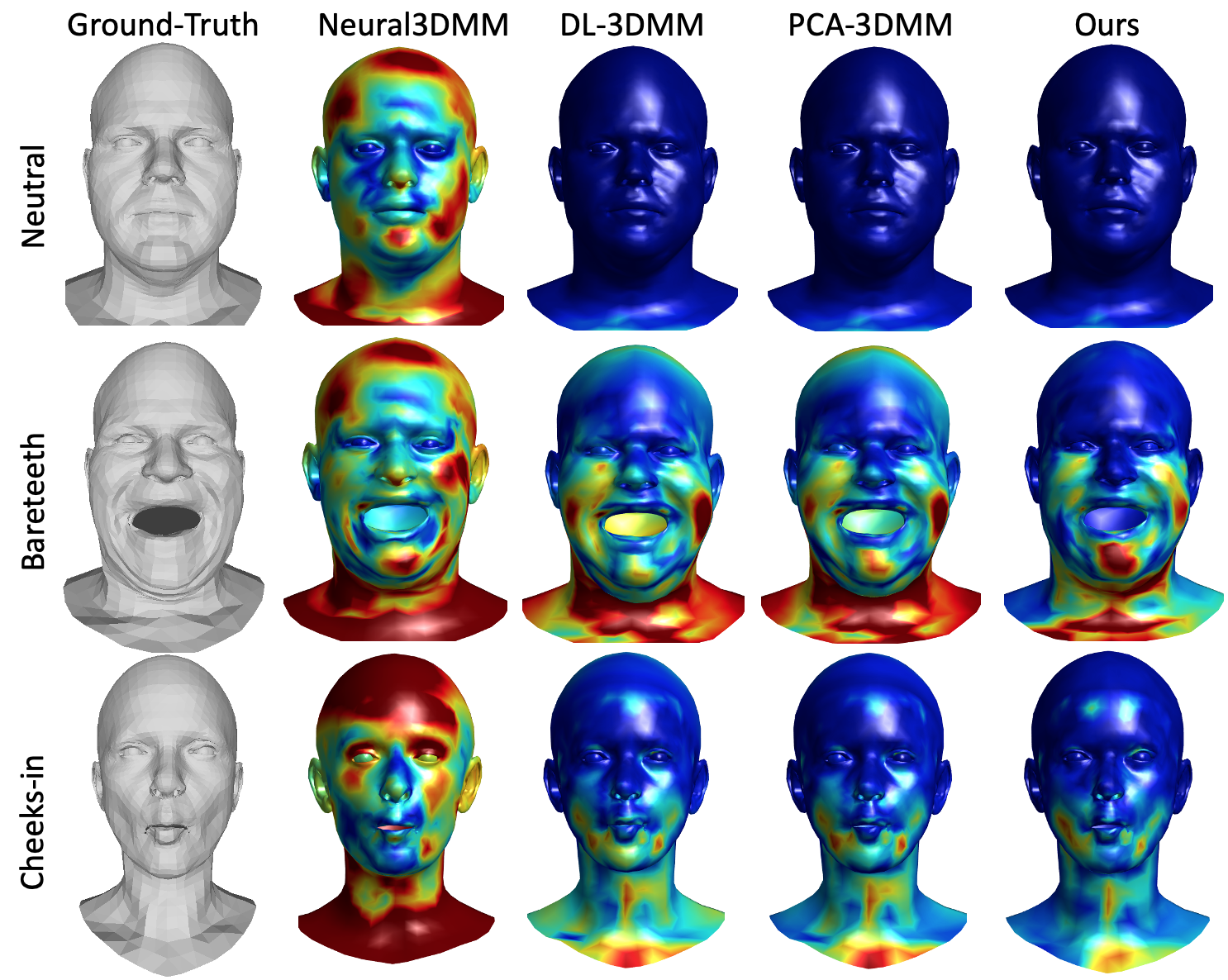}\\
\caption{Mesh reconstruction error (red=high, blue=low) of our model and other methods.}
\label{fig:cum-err}
\end{figure}

Figure~\ref{fig:cum-err} shows some qualitative examples by reporting error heatmaps in comparison with PCA, DL-3DMM~\cite{ferrari2017dictionary} and Neural3DMM~\cite{Bouritsas:ICCV2019} for the identity-independent splitting. The ability of our model as well as PCA and DL-3DMM to preserve the identity of the ground truth comes out clearly, in accordance with the results in Table~\ref{tab:comparison-ExprSplit}. By contrast, Neural3DMM shows high error even for the neutral faces, which proves its inability to keep the identity of an unseen face. Indeed, differently from to the other methods, Neural3DMM encodes the neutral face in a latent space and predicts the 3D coordinates of the points directly, which introduces some changes on the identity of the input face. This evidences the efficacy of our S2D-Dec, that instead learns per-point displacements instead of point coordinates.  


\subsubsection{Ablation Study}
We report here an ablation study to highlight the contribution of each loss used to train S2D-Dec, with particular focus on our proposed weighted-$L1$ reconstruction loss. We conducted this study on the CoMA dataset using the first three identities as a testing set and training on the rest. This evaluation is based on the mean per-vertex error between the generated and the ground truth meshes. We evaluated three baselines, $S1$, $S2$ and $S3$. For the first baseline ($S1$), we trained the model with the displacement reconstruction loss in~\eqref{eq:displacement_loss} only. In $S2$, we added the standard $L1$ loss to $S1$, which corresponds to our loss in~\eqref{eq:L1_weighted} without the landmark distance weights. To showcase the importance of weighting the contribution of each vertex, in $S3$ we added the landmark distance weights to the $L_{pr}$ loss. 
Results are shown in Table~\ref{tab:Ablation-study}, where the remarkable improvement of our proposed loss against the standard $L1$ turns out evidently. This is explained by the fact that assigning a greater weight to movable face parts allows the network to focus on regions that are subject to strong facial motions, ultimately resulting in realistic samples. 

\begin{table}[!t]
\centering
\small
\begin{tabular}{@{}lc@{}}
\toprule
Method & Error (mm) \\
\midrule
$S_1$ : $L_{dr}$ & $1.27 \pm 1.88$\\
$S_2$ : $S_1  + L_{pr}$ w/o distance weights & $0.92 \pm 1.33$ \\
$S_3$ :  $S_1 + L_{pr} $  & $0.50\pm 0.56$\\
\bottomrule
\end{tabular}
\caption{Ablation study on the reconstruction loss of S2D-Dec.}
\label{tab:Ablation-study}
\end{table}

\subsection{4D Facial Expression Evaluation}
Since Motion3DGAN generates samples from noise to encourage diversity, the generated landmarks and meshes slightly change at each forward pass. Thus, computing the mean per-vertex error with respect to ground-truth shapes as done in~\cite{PotamiasECCV2020} cannot represent a good and reliable measure in this case. So, we evaluated the quality of the generated expression sequences implementing an expression classification solution, similar to~\cite{PotamiasECCV2020}. We trained a classifier with one LSTM layer followed by a fully connected layer to recognize the $12$ dynamic facial expressions of CoMA given a landmarks sequence as input. We trained this classifier on the same sequences used to train Motion3DGAN. The first expression sample of each identity form the test set, resulting in $144$ testing samples. 

Since neither the dataset nor the code in~\cite{PotamiasECCV2020} are available for comparison, based on the information therein, we implemented a similar architecture relying on an LSTM to generate the per-frame expression and use it as baseline. The LSTM is trained to generate the motion of landmarks from an input code indicating the temporal evolution of the expression from neutral to the apex phase. Also this model was trained on the same data used for Motion3DGAN. 


For testing, we generated $144$ sequences with both Motion3DGAN and the LSTM generator. The sequences are consistent with those of the Motion3DGAN test set described above. These are then used to generate their corresponding meshes with our S2D-Dec. In Table~\ref{tab:comparison-lstm}, we report results in terms of classification accuracy and Frechet Inception Distance (FID)~\cite{heusel2017gans}. The metrics are computed either using the landmark sequences directly generated by Motion3DGAN and LSTM (Gen-LM row), or those extracted from the generated meshes (Det-LM row). Given that both Motion3DGAN and LSTM act as landmark generators, we also report the results obtained using sequences coming from a ``perfect'' generator, that is the ground truth landmark sequences of the test set (GT lands. column). This represents a sort of upper bound for the classification accuracy. We note that the features used to compute the FID metric are extracted by the last fully connected layer of our trained classifier, which outputs $512$ features per sequence.

 
In Table~\ref{tab:comparison-lstm}, we observe that, in all the cases, Motion3DGAN surpasses the accuracy of LSTM to a large extent, providing a clear evidence that the generated sequences better capture the expression dynamics. This is also supported by the lower FID, which indicates that the Motion3DGAN samples better approximate the ground truth motions. The same conclusion is drawn from the closer recognition rate of Motion3DGAN to that obtained with the ground truth sequences. Furthermore, the accuracy increases by first generating the corresponding meshes and then re-extracting the landmarks from them. This suggests the S2D-Dec is capable of maintaining the particular motion, which is also supported by the similar recognition rate obtained with ground truth landmarks ($73\%$) and those detected on their corresponding meshes generated with S2D-Dec ($73.61\%$).


\begin{table}[!t]
\centering
\small
\begin{adjustbox}{width=\columnwidth, center}
\begin{tabular}{@{}lccc|cc@{}}
\toprule
& \multicolumn{3}{c}{Classification Accuracy (\%) $\uparrow$} & \multicolumn{2}{c}{FID $\downarrow$}\\
\midrule
Method & GT lands& Mo3DGAN & LSTM & Mo3DGAN & LSTM\\
\midrule
Gen-LM & $\mathbf{73.00}$ & $\mathbf{65.28}$& $46.53$ & $\mathbf{20.45}$ & $21.76$\\
Det-LM & $\mathbf{73.61}$ & $\mathbf{69.44}$& $52.08$ & $\mathbf{19.01}$ & $27.96$\\
\bottomrule
\end{tabular}
\end{adjustbox}
\caption{Classification accuracy (\%) and Frechet Inception Distance (FID) obtained with Ground Truth (GT) landmarks, Motion3DGAN and LSTM. Results are obtained using either the generated landmarks directly (Gen-LM), or by extracting landmarks from the meshes resulting from applying the S2D-Dec to the landmarks motion (Det-LM).}
\label{tab:comparison-lstm}
\vspace{-0.2cm}
\end{table}


\begin{figure*}[!t]
\centering
\includegraphics[width=0.8\linewidth]{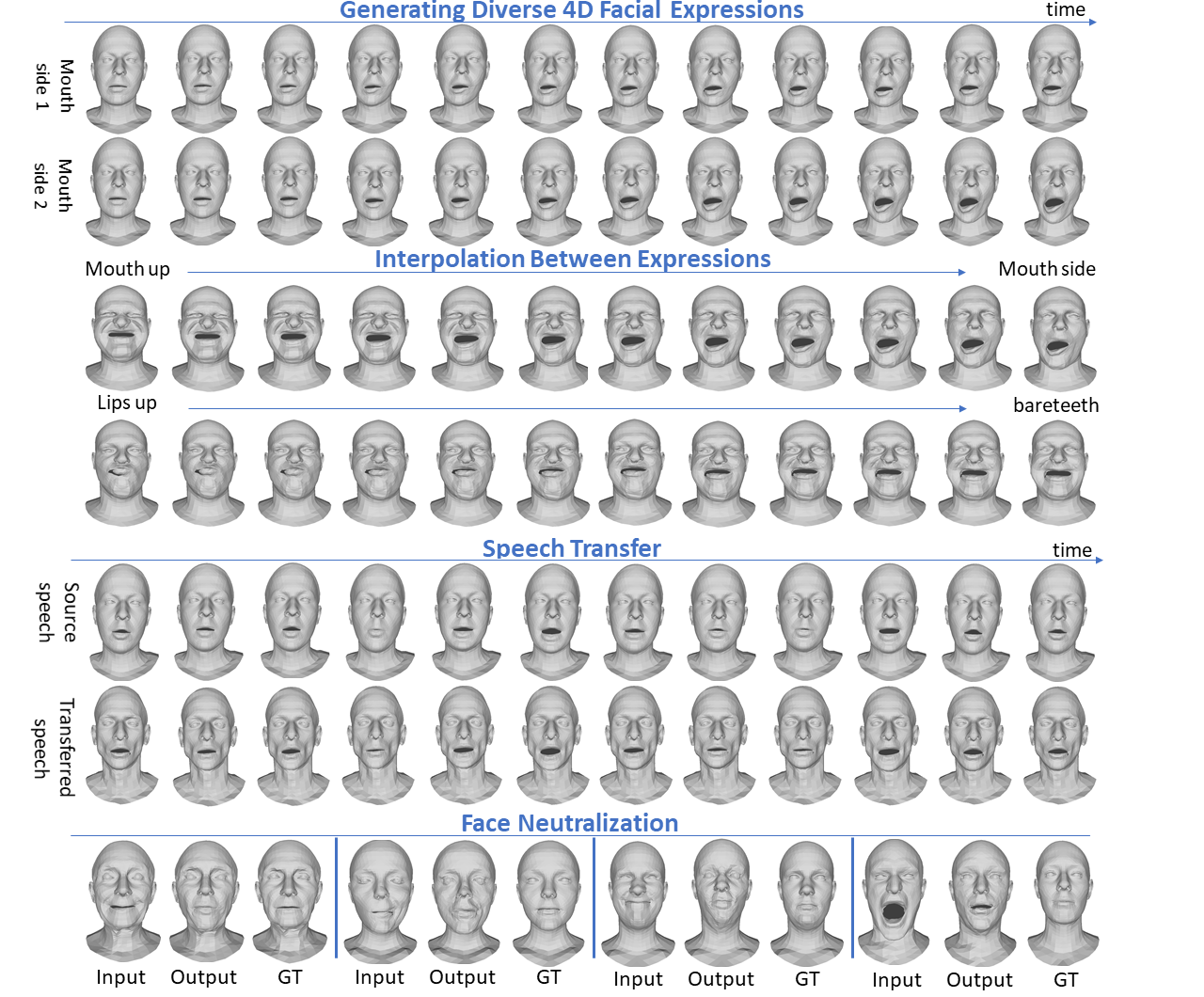}
\caption{\textbf{Applications} -- From top to bottom: \textbf{Diversity}: the same identity performing the same class of facial expression (mouth side) with two different motions generated by Motion3DGAN. \textbf{Interpolation}: dynamic expressions resulting from the interpolation between two expressions peaks. \textbf{Transfer}: speech transfer from one identity to another. \textbf{Face neutralization}: for each of the four examples, we show the input expressive face, the neutralized face with S2D-Dec, and the ground truth neutral face of the given identity.}
\label{fig:applications}
\end{figure*}

\subsection{Applications}
Our solution has some nice properties that open the way to various applications as shown in Figure~\ref{fig:applications}.

\noindent
\textbf{4D facial expression generation}: In the top two rows of Figure~\ref{fig:applications}, we show the ability of Motion3DGAN to generate sequences for the same expression label that are highly variegated. In spite of that, S2D-Dec is able to generalize and reconstruct realistic meshes.

\noindent
\textbf{Interpolation between facial expressions}: One interesting property of our Motion3DGAN is the possibility, enabled by the SRVF representation, of interpolating between generated motions. Given two points on the sphere $q_1$ and $q_2$, representing the motion sequences of two expressions, the geodesic path $\psi(\tau)$ between them is given by, $\psi(\tau)=\frac{1}{sin(\theta)}(sin(1-\tau)\theta) q_1 + sin(\theta\tau)q_2$, where, $\theta=d_{\mathcal{C}}\left(q_{1}, q_{2}\right)=\cos ^{-1}\left(\left\langle q_{1}, q_{2}\right\rangle\right)$. This path determines all the points $q_i$ existing between $q_1$ and $q_2$, each of them corresponding to a sequence of landmarks. Using our S2D-Dec, we can transform them to a 4D facial expression. 
Furthermore, while Motion3DGAN generates only neutral to apex sequences, we can exploit this interpolation to generate mixed 4D facial expressions that switch between the apex phase of different expressions by considering the last frame of each interpolated sequence. 
Figure~\ref{fig:applications}  illustrates the interpolated faces between the apex frames of two expressions (examples are given for Mouth-up to Mouth-side, and for Lips-up to Bare-teeth). 

\noindent
\textbf{Facial expression and speech transfer}: By using landmarks, our S2D-Dec can transfer facial expressions or speech between identities. This is done by extracting the sequence of landmarks from the source face, encoding their motion as an SRVF representation, transferring this motion to the neutral landmarks of the target face and using S2D-Dec to get the target identity following the motion of the first one. Some examples of speech transfer on the VOCASET dataset~\cite{VOCA2019} are shown in Figure~\ref{fig:applications} (see the supplementary material for their corresponding animations).

\noindent
\textbf{Neutralization}: Given an input expressive face, S2D-Dec can generate the corresponding neutral face. This is obtained by introducing the displacements between the landmarks of an expressive face and those of a neutral template to S2D-Dec, so to generate the displacements needed to neutralize the expression. The last row of Figure~\ref{fig:applications} shows that our model can neutralize the expressions to a great extent, even though such motions do not occur at all in the training data.









\section{Acknowledgments}
This work was supported by the French State, managed by National Agency for Research (ANR) National Agency for Research (ANR) under the Investments for the future program with reference ANR-16-IDEX-0004 ULNE and by the ANR project Human4D ANR-19-CE23-0020.  This paper was also partially supported by European Union's Horizon 2020 research and innovation program under grant number 951911 - AI4Media. 

\section{Conclusions and Limitations}
In this paper, we proposed a novel framework for dynamic 3D expression generation from an expression label, where two decoupled networks separately address modeling the motion dynamics and generating an expressive 3D face from a neutral one. We demonstrated the improvement with respect to previous solutions, and showed that using landmarks is effective in modeling the motion of expressions and the generation of 3D meshes. We also identified two main limitations: first, our S2D-Dec generates expression-specific deformations, and so cannot model identities. Moreover, while Motion3DGAN can generate diverse expressions and allows interpolating on the sphere to obtain complex facial expressions, the samples are of a fixed length (\ie, $30$ meshes, from neutral to apex). However, as shown in the applications, S2D-Dec can deal with motion of any length since it is independent from Motion3DGAN.



\section{Appendix}


\subsection{Landmarks Configuration}
\begin{figure}[!ht]
\centering
\includegraphics[width=0.7\linewidth]{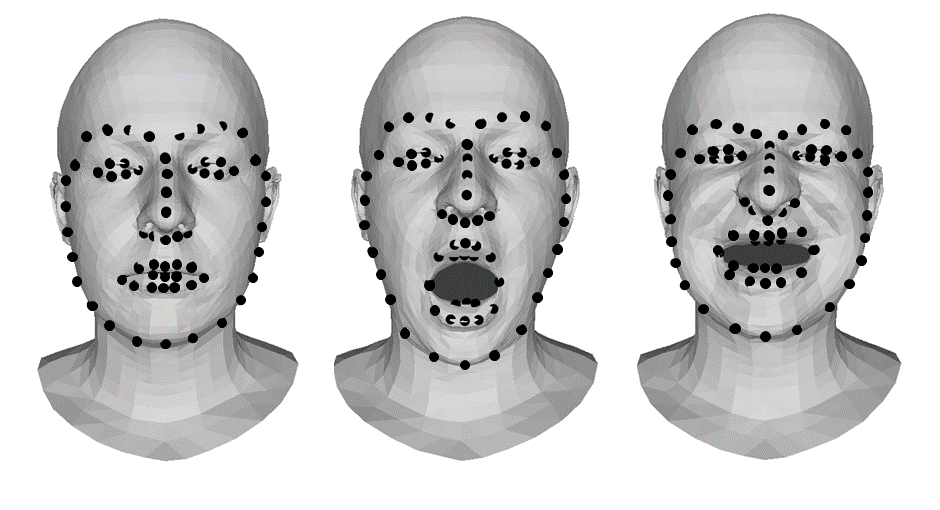}
\centering
\caption{Landmarks configuration used to guide our model.}
\label{fig:landmarks}
\end{figure}

In Figure~\ref{fig:landmarks} we show, for three different expressions, the configuration of landmarks used to guide the generation of the facial expression.

\subsection{Logarithm and Exponential Maps}
In order to map  the SRVF data forth and
back to a tangent space of $\mathcal{C}$, we use the logarithm  $\log _{p}(.)$ and the exponential $\exp _{p}(.)$ maps defined in a given point $p$ by,
\begin{equation}
\begin{split}
\label{Eq:LogSphereLog}
\log _{p}(q) &= \frac{d_{\mathcal{C}}(q,p)}{sin(d_{\mathcal{C}}(q,p))} (q - cos(d_{\mathcal{C}}(q,p))p), \\
\exp _{p}(s)&=cos(\|s\|)p + sin(\|s\|)\frac{s}{\|s\|},
\end{split}
\end{equation}

\noindent
where $d_{\mathcal{C}}(q,p)=\cos^{-1}(\langle q,p \rangle)$ is the distance between $q$ and $p$ in $\mathcal{C}$.

\subsection{Architecture of S2D-Dec}
The architecture adopted for S2D-Dec is based on the architecture proposed in~\cite{Bouritsas:ICCV2019}. S2D-Dec takes as input the displacements of $68$ landmarks illustrated in Figure~\ref{fig:landmarks}. The architecture includes a fully connected layer of size $2688$, five spiral convolution layers of $64$, $32$, $32$, $16$ and $3$ filters. Each spiral convolution layer is followed by an up-sampling by a factor of $4$. 

\section{Ablation Study}

\begin{figure}[!ht]
\centering
\includegraphics[width=\linewidth]{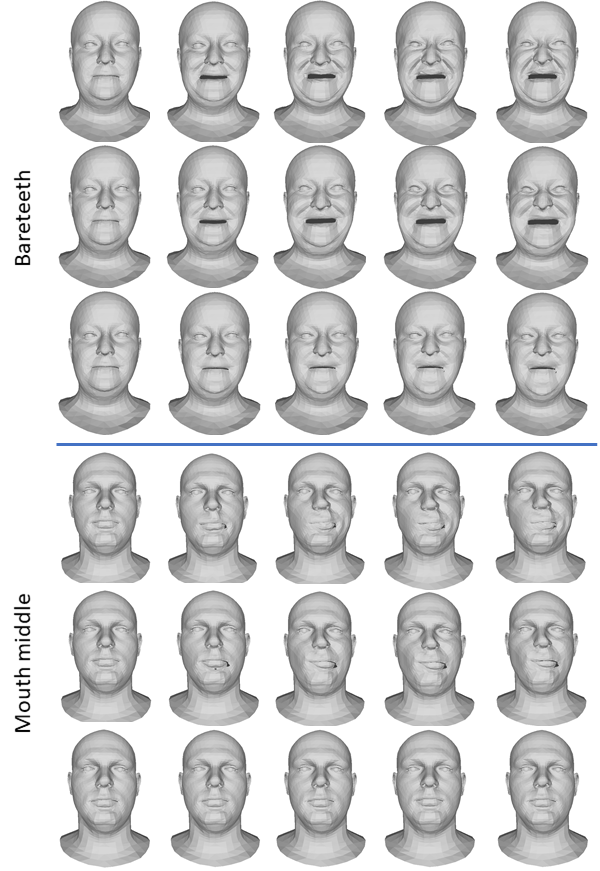}
\centering
\caption{Ablation study: qualitative comparison between ground truth (first row) our
model with (second row) and without (last row) weighted loss.}
\label{fig:ablationsupp}
\end{figure}

In this section, we report a visual comparison between reconstructions obtained with the standard L1 loss and our proposed weighted L1. Figure~\ref{fig:ablationsupp} clearly shows the effect of our introduced weighting scheme that allows for improved expression modeling. 

\end{document}